\pdfoutput=1

\documentclass[11pt]{article}

\usepackage{acl}

\usepackage{times}
\usepackage{latexsym}

\usepackage[T1]{fontenc}

\usepackage[utf8]{inputenc}

\usepackage{microtype}

\usepackage[utf8]{inputenc} 
\usepackage[T1]{fontenc}    
\usepackage{hyperref}       
\usepackage{url}            
\usepackage{booktabs}       
\usepackage{amsfonts}       
\usepackage{nicefrac}       
\usepackage{lipsum}
\usepackage{graphicx}
\graphicspath{ {./images/} }

\usepackage{soul}
\usepackage{url}
\usepackage[utf8]{inputenc}
\usepackage{amsmath}
\usepackage{booktabs}
\usepackage{algorithm}
\usepackage{algorithmic}
\usepackage[shortlabels]{enumitem}
\usepackage[TABBOTCAP]{subfigure}
\usepackage{multirow}
\usepackage{diagbox}

\usepackage{color}

\usepackage{xcolor}
\usepackage{soul}
\newcommand{\hlc}[2][yellow]{{%
    \colorlet{foo}{#1}%
    \sethlcolor{foo}\hl{#2}}%
}

\definecolor{babyblueeyes}{rgb}{0.63, 0.79, 0.95}
\newcommand{\Red}[1]{\textcolor[rgb]{1.00,0.00,0.00}{#1}}
\newcommand{\Blue}[1]{\textcolor[rgb]{0.00,0.00,1.00}{#1}}

\newcommand{\name}{CoCo\textsc{lm}}

\title{\name: Complex Commonsense Enhanced Language Model \\ with Discourse Relations}

\author{Changlong Yu\textsuperscript{$1$}\thanks{~~~Equal contribution.} \quad Hongming Zhang\textsuperscript{$1,2$}\footnotemark[1]\quad Yangqiu Song\textsuperscript{$1$}\quad Wilfred Ng\textsuperscript{$1$} \\
\textsuperscript{$1$}HKUST, Hong Kong, China \quad \textsuperscript{$2$}Tencent AI Lab, Seattle, U.S. \\
\texttt{\{cyuaq, yqsong, wilfred\}@cse.ust.hk},
\ \texttt{hongmzhang@tencent.com} \\ 

}

\begin{document}
\maketitle
\begin{abstract}

Large-scale pre-trained language models have demonstrated strong knowledge representation ability.
However, recent studies suggest that even though these giant models contain rich simple commonsense knowledge~(e.g., bird can fly and fish can swim.), they often struggle with complex commonsense knowledge that involves multiple eventualities~(verb-centric phrases, e.g., identifying the relationship between ``Jim yells at Bob'' and ``Bob is upset'').
To address this issue, in this paper, we propose to help pre-trained language models better incorporate 
complex commonsense knowledge.
Unlike direct fine-tuning approaches, we do not focus on a specific task and instead propose a general language model named \name.
Through the careful training over a large-scale eventuality knowledge graph \textsc{ASER}, we successfully teach pre-trained language models (i.e., BERT and RoBERTa) rich discourse-level commonsense knowledge among eventualities.
Experiments on multiple commonsense tasks that require the correct understanding of eventualities demonstrate the effectiveness of \name.

\end{abstract}

\section{Introduction}\label{sec:intro}

Recently, large-scale pre-trained language representation models (LMs)~\cite{devlin-etal-2019-bert,liu2019roberta} have demonstrated the strong ability to discover useful linguistic properties of syntax and remember an impressive amount of knowledge with self-supervised training over a large unlabeled corpus~\cite{DBLP:conf/emnlp/PetroniRRLBWM19,jiang-etal-2020-know}.
On top of that, with the help of the fine-tuning step, LMs can learn how to use the memorized knowledge for different tasks, and thus achieve outstanding performance on many downstream natural language processing~(NLP) tasks.

\begin{table}[h]
\small
    \centering
    \begin{tabular}{p{5.6cm}|p{1.1cm}}
    \toprule
        Query & Answer \\
    \midrule
        Birds can [MASK]. & \Blue{fly} \\
        Cars are used for [MASK]. & \Blue{transport} \\
        \midrule
        Jim yells at Bob, [MASK] Jim is upset. & \Red{but}\\
        Jim yells at Bob, [MASK] Bob is upset. & \Red{but} \\
    \bottomrule
    \end{tabular}
    \vspace{0.1in}
    \caption{Exploring knowledge contained in pre-trained language models following LAMA~\cite{DBLP:conf/emnlp/PetroniRRLBWM19}. Queries and prediction returned by BERT-large are presented. Semantically plausible and implausible prediction are indicated with \Blue{blue} and \Red{red} colors. }
    \vspace{-0.1in}
    \label{tab:intro_example}
\end{table}

As discussed in~\citet{DBLP:journals/corr/abs-2007-00849}, while language models have already captured rich knowledge, they often only perform well when the semantic unit is a single token while poorly when the semantic unit is more complex (e.g., a multi-token named entity or an \textbf{eventuality}, which is a linguistic term for verb-centric phrases covering \textit{activities}, \textit{states} and \textit{events}~\cite{bach1986algebra,araki-mitamura-2018-open}).
For example, as shown in Table~\ref{tab:intro_example}, if we follow LAMA~\cite{DBLP:conf/emnlp/PetroniRRLBWM19} to analyze the knowledge contained in BERT-large~\cite{devlin-etal-2019-bert} with a token prediction task, we can find out that BERT can understand that birds can fly, and a car is used for transportation, but it fails to understand the relation between ``Jim yells at Bob'' and relevant eventualities.
An important reason behind this is that current language models heavily rely on token-level masked language models (MLMs) as the loss function, which can effectively represent and memorize token co-occurrence statistics\footnote{~\citet{sinha-etal-2021-masked} also explains the success of LMs due to distributional information. These models pretrained over sentences with shuffled word order still achieve high accuracy.} but struggle at perceiving multi-token concepts.

\begin{table*}[t]
\small
  \centering
  \begin{tabular}{p{1.0cm}|p{14.0cm}}
  \toprule
     Type  &  Sequences \\
  \midrule
  Temporal & I had a dream. \hlc[pink]{\textit{Precedence}} (\textbf{Before}) I met with you yesterday. \hlc[pink]{\textit{Succession}} (\textbf{After}) There were so many matters.  \\
  \midrule
  Casual &   I go to supermarket. \hlc[pink]{\textit{Reason}} (\textbf{Because}) I have a coupon. \hlc[pink]{\textit{Result}} (\textbf{So}) The price is great. \\ 
  \midrule
  Others & You can come with me. \hlc[pink]{\textit{Alternative}} (\textbf{Or}) You can stand here. \hlc[pink]{\textit{Contrast}} (\textbf{But}) The situation remains unchanged.  \\
  \bottomrule
  \end{tabular}
  \vspace{0.1in}
  \caption{Examples of eventuality sequences with different types of discourse relations~(highlighted with pink) and connectives~(bolded). Note there may exist multi-relational eventuality pairs}\label{tab:seq_examples}
  \vskip -1em
\end{table*}

To address this problem and equip LMs with complex and accurate human knowledge, several recent works attempt to integrate entity representations from external knowledge graphs.  
While those approaches have been proved effective in merging structured knowledge into the LMs, they still have two limitations when applying to eventuality representations: 
(1)~The first line of work~\cite{DBLP:journals/corr/abs-2007-00849,shen-etal-2020-exploiting,fevry2020entities} restricts a fixed set of named entities or concepts to be linked to KGs while the eventualities are not easily canonicalized.
There are enormous eventualities, which many of them refer to similar meanings such as ``Tom is upset'' and ``Alice is upset''.
(2)~The second class of methods uses powerful contextualized representations to encode one-hop triplets from KGs~\cite{bosselut2019comet,yao2019kg} for the task of KG completion.
However, it is not sufficient for tasks that require the understanding of complex discourse relations in the event sequences or chains.   
For example pretrained LMs on the story ending prediction task~\cite{mostafazadeh-etal-2016-corpus} have gaps with human performance~\cite{li2019story}.
Besides that, different types of relations~(\textit{casual} or \textit{temporal}) make high-order inference over eventualities difficult and challenging.

In this paper, to effectively inject eventuality knowledge into pre-trained language representation models, we propose a knowledge injection framework \name, which requires no concept or eventuality linking and encodes multi-hop eventuality information as well as their discourse relations.
The starting point is a large-scale eventuality knowledge graph, \textsc{ASER}~\cite{zhang2020aser}, where the edges are discourse relations among eventualities~(e.g., ``being hungry'' can cause ``eat food'' and ``being hungry'' often happens at the same time as ``being tired'').
First, we go beyond one-hop modeling~\cite{yao2019kg,bosselut2019comet} and carefully conduct weighted random walk over \textsc{ASER} to harvest multi-hop eventuality sequences connected by discourse relations~($\S$\ref{sec:method_sequence}).
Individual eventualities are contextualized by coherent sequences~(examples in Table~\ref{tab:seq_examples}).
Second, we fine-tune pretrained LMs on the sampled sequences and reformulate the masked language modeling objective to new \textit{eventuality-level} masking to perceive the eventualities as independent semantic units~($\S$~\ref{sec:masking}).
In addition, two auxiliary tasks of discourse relation prediction are proposed to make implicit commonsense inferences~($\S$~\ref{sec:aux_tasks}).
For example, the new tasks explicitly reinforce the casual relation prediction between ``I have a coupon'' and ``The price is great''.
By doing so, we successfully expose and inject fruitful high-order information between eventualities to pretrained LMs.
To understand the impact of our proposed \name, we conduct experiments on three tasks that require the understanding of temporal, causal, mixed~(multiple) relations respectively.
The results show that our method achieves substantial improvements on the multiple-relation task while competitive performance on single-relation tasks.
Extensive analyses are conducted to show the effect and contribution of all components in \name.
Our main contributions are as follows:
\begin{itemize}[leftmargin=*]
    \item We propose CoCo\textsc{lm}, a new contextualized language model enhanced by complex commonsense knowledge from high-order discourse relations. 
    CoCo\textsc{lm} is trained to predict the whole eventuality among the sequences using a large-scale eventuality KG. 
    \item We introduce two auxiliary discourse tasks to help incorporate discourse-related knowledge into pre-trained language models, which complement the special eventuality masking strategy. 
    \item CoCo\textsc{lm} achieves stronger performance than the baseline LMs on multiple datasets that require the understanding of complex commonsense knowledge about eventualities.\footnote{Our code and models are available at \url{https://github.com/HKUST-KnowComp/Co2LM}.}
\end{itemize}






\section{Methods}\label{sec:Methods}

\begin{figure}[t]
  \centering
  \includegraphics[width=0.8\columnwidth]{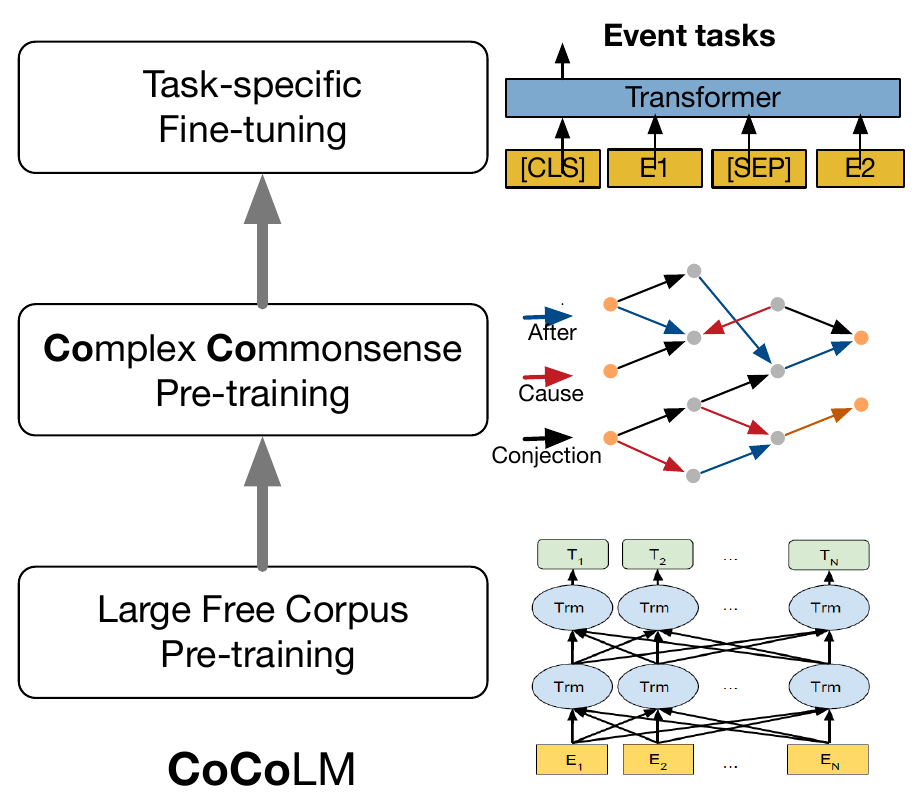}
  \caption{The overall framework of \name. On top of base pre-trained LMs, complex commonsense knowledge from the eventuality sequences is injected by fine-tuning MLMs and auxiliary discourse tasks.
  }\label{fig:framework}
  \vspace{-0.1in}
\end{figure}

The overall framework of \name ~is presented in Figure \ref{fig:framework}. 
Given a pre-trained language model, we inject complex commonsense knowledge about eventualities by adding one adaptive pre-training stage~\cite{Gururangan2020DontSP}.
Specifically, we first generate eventuality sequences based on carefully controlled walks over existing eventuality knowledge graphs and then use the sequences as the context to help LMs handle eventualities.
Besides the original token-level MLM objective, we also introduce the \textit{eventuality-level} masking strategy and several auxiliary tasks to assist the training.
As the training is not task-specific, the resulting LM can be easily applied to any downstream tasks via another task-specific fine-tuning stage.


\subsection{Eventuality Sequence Generation}\label{sec:method_sequence}


Multi-hop path information have been shown useful and interpretable to provide extra context knowledge by connecting the concepts from ConceptNet~\cite{speer2017conceptnet} for commonsense reasoning tasks~\cite{lin-etal-2019-kagnet,wang-etal-2020-connecting}.
Similarly, we generate eventuality sequences by leveraging \textsc{ASER}, which uses eventualities as nodes and the discourse relations as edges.
\textsc{ASER} extracts rich eventuality knowledge from diverse corpus, such as ``being hungry'' and ``being tired'' often happen together and people often ``make a call'' before they go.
It contains much larger scale of discourse relations than DisSent~\cite{nie-etal-2019-dissent}.
Interestingly, beyond the single edges, higher-order connections over \textsc{ASER} can also reflect insightful eventuality knowledge. For example, ``sleep'' and ``go'' are not likely to happen at the same time because ``sleep'' can be caused by ``being tired'' and there exist \textit{contrast} connections between ``being tired'' and ``go''.
To include higher-order knowledge into the model, we propose to take the whole graphical structure into consideration rather than single-hop edges.
Motivated by DeepWalk~\cite{DBLP:conf/kdd/PerozziAS14}, we randomly sample paths to simulate the overall graph structure and generate eventuality-level co-occurrence information.

Given the initial knowledge graph $\mathcal{G} = (\mathcal{E}, \mathcal{R})$, where $\mathcal{E}$ is the eventuality set and $\mathcal{R}$ is the relation set, we conduct the weighted random walk based on the edge weights over $\mathcal{G}$ to sample eventuality paths.
We denote each path as $(E_0, r_0, E_1, r_1,..., r_{l-1}, E_l)$, where $E$ means an eventuality, $r$ a discourse edge connecting two eventualities, and $l$ the numbers of eventualities along the sequence.
To convert the sampled sentence into a token list, we keep all words in each event as a sentence and use representative connectives for each discourse relation to connect them~(examples in the Table~\ref{tab:seq_examples}; full list in the Appendix Table~\ref{tab:relation_marker}).
As \textsc{ASER} is automatically extracted from raw corpus, it may contain noise.
To minimize the influence of the noise and improve the informativeness, the selected paths should fulfill:
\begin{enumerate}[leftmargin=*]
    \item To filter out rare eventualities, the frequency of starting eventualities has to be larger than five.
    \item Other than the relations that have the transitive property (e.g., \textit{Precedence}, \textit{Result}), each selected path should not contain successive edges with repeated relations. 
    \item We manually improve the sampling probability of selecting sub-sequence patterns like ``$E_i$ \textbf{Condition} $E_j$ \textbf{Reason} $E_k$''. Since it has been proven that 
    \textit{if}-\textit{then} rules~\cite{sap2019atomic} and  \textit{if}-\textit{then}-\textit{because} rules~\cite{arabshahi2020conversational} are crucial for reasoning. 
\end{enumerate}

   

\subsection{Eventuality-Level Mask}\label{sec:masking}



\begin{figure*}[ht]
  \centering
  \includegraphics[width=2.0\columnwidth]{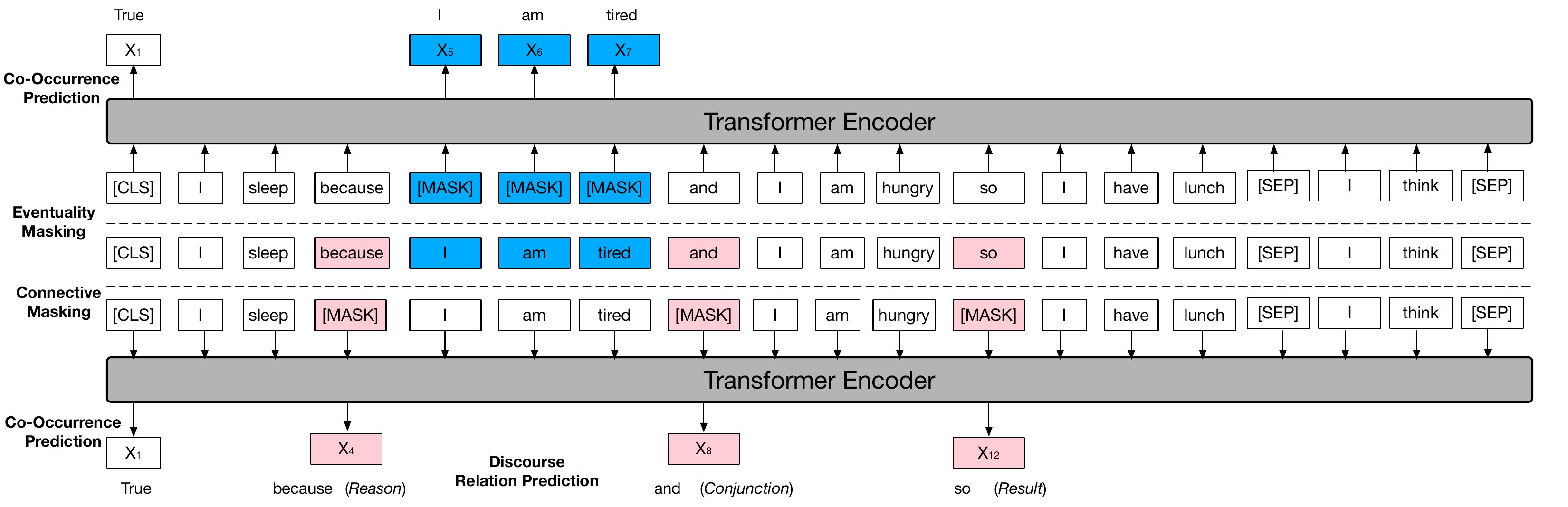}
  \caption{Illustration of \name -(Complex commonsense pre-training stage). Given an eventuality sequence, it is either masked by the whole eventuality masking~(\texttt{in blue}) or discourse connective masking strategy~(\texttt{in pink}). 
  Besides the regular masked language model, the discourse relation labels are jointly predicted for masked connective tokens (on $\mathbf{x}_4$, $\mathbf{x}_8$ and $\mathbf{x}_{12}$). 
  Co-occurrence prediction (on $\mathbf{x}_1$) is conducted for both masking strategies. 
  }\label{fig:method}
\end{figure*}

Masking strategy plays a crucial role in the training of language representation models.
Besides the random token-level masking strategy, many other masking strategies have been explored by previous literature such as the-whole-word masking~\cite{devlin-etal-2019-bert,cui2019pre}, entity masking~\cite{sun2019ernie,shen-etal-2020-exploiting} or text span masking~\cite{joshi2020spanbert}\footnote{Unlike \texttt{SpanBERT}, we have discourse connectives as span boundaries and do not need the SBO objective.}.
Similarly, to effectively help the model view each eventuality as an independent semantic unit, we propose the following two masking strategies: (1) \textbf{Whole Eventuality Masking}: Similar to the whole word masking or entity masking strategies, the whole eventuality masking aims to reduce the prior biases of eventuality tokens. 
For example, given an eventuality sequence ``I
feel sleepy because I drink a cup of [MASK].'', BERT would easily predict ``coffee'' or ``tea'' because of the prior knowledge of ``cup of'' inside the eventuality. 
Instead of that, masking the whole ``I drink a cup of coffee'' would encourage the prediction to treat each eventuality as an independent semantic unit and focus on the relations between them. 
For each sampled sequence, we randomly mask at most one eventuality to fulfill the masking budget, which is typically 25\% of the sequence token length.
(2) \textbf{Discourse Connective Masking}: 
Besides masking the eventualities, to effectively encode the discourse information, we also tried masking the discourse connectives.

Examples of two masking strategies are shown in Figure~\ref{fig:method}.
It is worth mentioning that for each sequence, we only randomly select one type of masking strategy to guarantee that enough information is kept in the left tokens for the prediction.
The formal masking objective is defined as follows. Given a tokenized sampled sequence $X=(x_1, x_2, ..., x_n)$, 
after masking several tokens, we pass it to a transformer encoder~\cite{vaswani2017attention} and denote the resulting vector representations as $\mathbf{x_1}, \mathbf{x_2}, ... \mathbf{x_n}$.
The training loss $\mathcal{L}_{mlm}$ can thus be defined as:
\begin{equation}
    \mathcal{L}_{mlm} = -\frac{1}{|\mathbf{M}|}\sum_{i \in \mathbf{M}}{\rm log} P(x_i | \mathbf{x_i}),
\end{equation}
where $\textbf{M}$ means the set of masked tokens following the aforementioned masking strategies.

\subsection{Auxiliary Tasks}\label{sec:aux_tasks}

A limitation of the MLM loss is that the prediction is over the entire vocabulary, and as a result, the model could not effectively learn the connection between eventualities and connective words. 
To remedy this and force the model to learn the discourse knowledge, we propose to add an additional classification layer after the last layer of transformer encoder and it feeds the output vector $\mathbf{x_i}$ of connective token $x_i$ into a softmax layer over the set of discourse relation labels as follows.

\begin{equation}
    P(l_i | \mathbf{x_i} ) = {\rm softmax}(\mathbf{x_i}\mathbf{W} +\mathbf{b}),
\end{equation}
\begin{equation}
    \mathcal{L}_{rel} = -\sum_{i \in \mathbf{M}_{R}} \log P(l_i = \widetilde{l}_i | \mathbf{x}_{i}),
\end{equation}
where $\mathbf{M}_{R}$ is the index set of masked discourse connective tokens (e.g., \textit{because}, \textit{and}, \textit{so}) in Figure~\ref{fig:method}, $l_i$ is the predicted discourse relation label, and $\widetilde{l}_i$ the label provided by \textsc{ASER}~(10 relations in Table~\ref{tab:relation_marker}).
$\mathbf{W}$ and $\mathbf{b}$ are trainable parameters. 

Besides the aforementioned discourse relations, \textsc{ASER} also provides the \textit{Co-occurrence} relations between eventualities, which mean that two eventualities appear in the same sentence, but there are no explicit discourse markers between them.
Co-occurrence information has been used for  narrative event prediction~\cite{chambers-jurafsky-2008-unsupervised}
Though \textit{Co-occurrence} relations are less informative, high frequency pairs still reflect rich knowledge about eventualities.
Motivated by this, we propose another auxiliary task to help the model to learn such knowledge.
Specially, given an eventuality sequence $S= (E_0, r_0, E_1, r_1, ..., r_{l-1}, E_{l})$ and an eventuality $E_c$, we format the input\footnote{The special tokens are based on the base model, i.e., we add ``[CLS]'' and ``[SEP]'' for \texttt{BERT} models and add ``<s>'' and ``</s>'' for \texttt{RoBERTa} models. All notations in the rest of this paragraph are based on \texttt{BERT}.} as ``[CLS] $S$ [SEP] $E_c$ [SEP]''. 
We set 50\% of the time $E_c$ to be the positive \textit{co-occurred} eventuality with one of the eventualities in the sequence while 50\% of the time $E_c$ is randomly sampled negative in \textsc{ASER}.
Similar to the next sentence prediction in the original BERT, on top of the vector representation of token [CLS], i.e., $\mathbf{x}_{cls}$, we add another classification layer to predict whether the \textit{Co-occurrence} relations hold or not.
The training objective $\mathcal{L}_{occur}$ for binary classification is similar to $\mathcal{L}_{rel}$: 
\begin{equation}
    \mathcal{L}_{occur} = -\log P(l_i=  \widetilde{l}_i, |\mathbf{x}_{cls}),
\end{equation}
where $\widetilde{l}_i$ is the true co-occurrence label~(positive or negative) for the sequence. 

Merging all three losses together, we can then define the overall loss function $\mathcal{L}$ as:
\begin{equation}
    \mathcal{L} = \mathcal{L}_{mlm} + \mathcal{L}_{rel} + \mathcal{L}_{occur}.
\end{equation}

\section{Experiments}

\subsection{Implementation Details}\label{sec:implementations}

In this work, we use the released \textsc{ASER}-core version\footnote{\url{https://github.com/HKUST-KnowComp/ASER}} extracted from multi-domain corpora, which contains over 27.6 million eventualities and 8.8 million relations. 
We follow the heuristic rules in Sec.~\ref{sec:method_sequence} to sample eventuality sequences for pre-training. 
Overall we generated 4,041,572 eventuality sequences~(sentences), ranging from one to five hops and the one-hop sequence means the direct~(first-order) edge in the \textsc{ASER}.
We also down-sample eventuality nodes with extremely high frequency such as \textit{I see}.
Sequence examples are listed in Table~\ref{tab:seq_examples} and more examples as well as sequence distribution over different lengths are appended in the Appendix. 


We select base and large version of \texttt{BERT}~\cite{devlin-etal-2019-bert}, \texttt{RoBERTa}~\cite{liu2019roberta} as the base LM.
For the continual complex commonsense pre-training phase, we use the Adam optimizer for 10 epochs with batch size 128, learning rate 1e-5 and weight decay 0.01.
Considering the relative longer span of masked eventualities, we enlarge the masking proportion from 15\% to 25\%, which averagely add 1.7 more masked tokens in the sequences. 
We implemented the pretraining with Huggingface library~\cite{wolf-etal-2020-transformers} and running \name ~pretraining on eight Nvidia V100 32GB GPUs took four days. 
Pretraining introduces two classification layers with thousands of parameters.

\subsection{Datasets and Evaluations}
\label{sec:experiments}

In this section, we introduce evaluation datasets and settings as follows:

\noindent \textbf{ROCStories}~\cite{mostafazadeh-etal-2016-corpus} is widely used for story comprehension tasks such as Story Cloze Test. 
It contains 98,162 five-sentence coherent stories as the unlabeled training dataset, 1,872 four-sentence story contexts along with two candidate ending sentences in the dev and test datasets.  
The dataset split for the story ending prediction task is the same as \citet{li2019story}.

\noindent \textbf{MATRES}~\cite{ning-etal-2018-multi} is a pairwise event temporal ordering dataset, where each event pair in one document is annotated with a temporal relation (\textit{Before, After, Equal, Vague}). 
It contains 13,577 event pairs extracted from 256  documents for training~(25 left for dev) and 20 for testing. 

\noindent \textbf{COPA}~\cite{gordon-etal-2012-semeval} is a binary-choice commonsense causal reasoning task, which requires models to predict which the candidate hypothesis is the plausible effect/cause of the given premise. 
We follow the training/dev/test split in SuperGLUE~\cite{wang2019superglue}.  

\begin{table}[t!]\small
\setlength\tabcolsep{4.0pt}
\centering
\begin{tabular}{p{1.5cm}|p{1.3cm}| p{1.0cm} |p{0.9cm} | p{0.9cm} }
\toprule
\textbf{Dataset}& \textbf{Type} & \textbf{\# Train} & \textbf{\# Dev} & \textbf{\# Test}\\
\midrule
ROCStories & Narrative (Multiple) & 1,771  & 100 & 1,871  
\\ \hline
MATRES  & Temporal & 231\textsuperscript{*} & 25 & 20 \\
 \hline
COPA   & Causal & 400 & 100 & 500 \\
\bottomrule
\end{tabular}
\caption{The statistics of evaluation datasets~(See examples in \ref{ap:dataset_examples}). The tasks are binary or multiple classification problems. Note the dataset of MATRES is split at the article level following~\citet{ballesteros2020severing}. \label{tab:dataset}}  
\end{table}

\begin{table*}[h]
\small
\centering
\setlength\tabcolsep{7.5pt}
\begin{tabular}{l||cc|cc|c}
\toprule
     & \multicolumn{2}{c}{{\bf ROCStories}}  & \multicolumn{2}{|c|}{{\bf MATRES}} & {\bf COPA} \\  \cmidrule(r){2-3} \cmidrule(lr){4-5} \cmidrule(l){6-6}
     Model  &  Accuracy  &  Accuracy (D) & Accuracy  & F1  & Accuracy   \\ \midrule
 BERT-base &  52.9  &  45.9  &  71.5  &  77.2  &  69.8   \\
 \name ~(\texttt{BERT-base}) &   84.2  &  65.2 &  72.8 &  77.8 &  73.8   \\\midrule
 BERT-large &  88.9 &  69.1 &  73.5 & 78.9 &  70.6   \\
 \name ~(\texttt{BERT-large}) &  91.9 &  71.2  &  73.9  &  79.2 &  75.8  \\ \midrule
 RoBERTa-base & 93.3 &   73.2 &  74.0  &  79.2 &   85.4  \\ 
 \name ~(\texttt{RoBERTa-base}) &  94.1 &  75.2&   74.2 &  79.8 &  86.2 \\ \midrule
 RoBERTa-large &  97.4&  88.1 &  75.2&  81.0 &  90.6\\
 \name ~(\texttt{RoBERTa-large}) &  \textbf{97.9}&  \textbf{89.4} &  \textbf{75.5} &  \textbf{81.6} &  \textbf{91.3} \\
 
\bottomrule

\end{tabular}
\caption{\label{tab:expresult} Evaluation results on three commonsense task (top scores in boldface). We report the accuracy of \texttt{ROCStories} dataset under normal supervised setting and debiased~(D) setting mentioned in the $\S$\ref{sec:experiments}.}  
\end{table*}

The statistics of the three selected datasets are presented in Table~\ref{tab:dataset}. 
For fine-tuning experiments, we select the learning rate from \{2e-5, 1e-5, 5e-6\}, and maximize the sequence length and batch size such that they can fit into GPU memory.
Fine-tuning was much faster due to fewer new parameters from classification layers.

Different from MATRES and COPA, solving the story ending tasks of ROCStories requires multi-type relation inferences including causal, temporal etc. 
Moreover, as mentioned in~\citet{DBLP:conf/acl/SharmaABM18}, there is a strong bias about the human-created negative endings such that the model can distinguish the positive and negative endings without seeing the first four events. 
Even though \citet{DBLP:conf/acl/SharmaABM18} tried to filter the annotations, the bias still cannot be fully relieved. 
As a result, to clearly show the effect of adding complex knowledge about events into the LM, besides the most widely used supervised setting, we also report the performance of a \textit{debiased setting}, where the model randomly selects events from other stories as the negative ending during the training.
The debiased setting is indicated with ``D''.
Following previous works, we report accuracy for the ROCStories, MATRES and COPA tasks. 
For MATRES, we also report F1 scores by considering the task as general relation extraction and treating the label of \textit{vague} as \textit{no relation}~\cite{ning-etal-2019-improved}. 
All models are trained until convergence and the best model on the dev set is selected to be evaluated.




\section{Experimental Results}\label{sec:results}



The results are presented in Table~\ref{tab:expresult}, from which we can see that CoCo\textsc{lm} consistently outperforms all the baselines on all three commonsense tasks, especially on the debiased setting of ROCStories.

Besides that, we can make the following observations. 
First, the improvement of our model is more significant on ROCStories than COPA and MATRES, which is mainly because multiple relation combinations in the eventuality sequences bring high-order information and thus help complex reasoning.
Second, \name ~achieves significant improvement on lower-capacity LMs trained on small corpora. For example, \name ~brings up to 59.2\% improvement over BERT-base on the ROCStories dataset. 
Third, compared with the original supervised setting, the debiased setting is more challenging for all models, which helps verify our assumptions that previous models might benefit from the bias. Here the debiased setting should be more fair for comparison. 
When we dig into the MATRES dataset, event pairs~(typically verb pairs) are associated with the context, which some of could be easily inferred from the local context with obvious clues~\cite{ballesteros2020severing} while the others may need external commonsense knowledge that can be memorized by the language models. As a comparison, both the ROCStories and COPA do not have any extra context, and thus require the pre-trained LMs to know the essential knowledge to solve those problems.

In the rest of this section, we conduct extensive experiments and case studies to demonstrate the contribution of different components. In all following analysis experiments, we use \texttt{BERT-large} as the base language model and \texttt{ROCStories} as the evaluation dataset.


\begin{table}[t]\small
\setlength\tabcolsep{1.3pt}
\centering
\begin{tabular}{l||cc|cc}
\toprule
\textbf{Method}& Accuracy & $\Delta$ & Accuracy (D) & $\Delta$ (D) \\
\midrule
CoCo\textsc{lm} & 91.9 & - & 71.2 & - \\
\midrule
\ w/o \textit{occur loss}  & 91.3 & -0.6 & 70.4 & -0.8 \\
\ w/o \textit{eventuality mask} & 91.1 & -0.8 & 70.2 & -1.0  \\
\ w/o \textit{rel loss}  & 90.5 & -1.4 & 69.6 & -1.6 \\ 
\ w/o \textit{occur} \& \textit{rel losses} & 90.3 & -1.6 & 69.3 & -1.9  \\ \midrule
\ w token-level \textit{mlm} only  & 89.2 & -2.7 & 69.2 & -2.0  \\
\bottomrule
\end{tabular}
\caption{Ablation study on ROCStories test set by removing different model components. \textit{occur} and \textit{rel} are discourse relation and co-occurrence loss respectively.\label{tab:ablition_method}}  
\end{table}

\subsection{Ablation Study}

We conduct an ablation study in Table~\ref{tab:ablition_method} via LM pretraining with different settings and then finetuning. 
We can see that all components contribute to the final success of our model, especially the \textit{Relation loss}. 
This result again verified that discourse connective prediction is a much more challenging pretraining task as shown in \citet{malmi2018automatic}.
\name ~is optimized to memorize high-order discourse knowledge that strongly correlates with downstream tasks and thus brings more performance boost. 
Besides, when replacing the whole eventuality masking with random token masking, we can observe 0.8\%~(1.0\%) accuracy drop, which indicates the usefulness of eventuality-level masking. 
The relative better improvement of \textit{Co-occurrence loss} suggests our previous assumption that even though compared with other discourse relations (e.g., \textit{Before} and \textit{Cause}), the \textit{Co-occurrence} relations have relatively weaker semantic, it still can help models to better understand events due to its large scale.

\begin{table}[t]\small
\setlength\tabcolsep{1.8pt}
\centering
\begin{tabular}{l||cc|cc}
\toprule
\textbf{Resource} & Accuracy & $\Delta$ & Accuracy (D) & $\Delta$ (D) \\
\midrule
ASER (M) & 91.9 & - & 71.2 & -  \\
\midrule
ASER (S) & 85.4 & -6.5 & 67.5 & -3.7 \\
ATOMIC (S) & 88.2 & -3.7 & 68.2 & - 3.0 \\
\bottomrule
\end{tabular}
\vspace{-0.1in}
\caption{Effect of different event knowledge resources. ``M'' and ``S'' pertain to ``multi-hop'' and ``single-hop''.\label{tab:ablation_resource}}  
\end{table}

\begin{table*}[t]\small
\setlength\tabcolsep{4.0pt}
\centering
\begin{tabular}{p{1.2cm} | p{8.0cm} | p{2.9cm} | p{2.9cm} }
\toprule
\textbf{Dataset} &\textbf{Example}  & \textbf{BERT [MASK]} & \textbf{\name} \textbf{[MASK]} \\
\midrule
ROC

Stories    & \underline{Context}: Ed made beef jerky for a living. He ran the business out of his garage. One day he woke up and noticed his garage jarred open. He looked inside and noticed everything in disarray

\underline{Positive Ending}: Ed was delighted to see this.

\underline{Negative Ending}: Ed was shocked called the police for an investigation. 
& Context + [MASK] + Ending:

\hlc[babyblueeyes]{\textbf{P: when, then, while}}

\hlc[babyblueeyes]{\textbf{N: and, but, so}}

& 
Context + [MASK] + Ending:

\hlc[pink]{\textbf{P: so, hence, therefore}}

\hlc[pink]{\textbf{N: or, and, though}}

\\ \hline
MATRES  &  The last surviving member of the team which first conquered Everest in 1953 has \underline{\{$e_1$: died\}} in a Derbyshire nursing home.

George Lowe, 89, \underline{\{$e_2$: died\}} in Ripley on Wednesday after a long-term illness, with his wife Mary by his side. &S1, +[MASK] + S2:

\hlc[babyblueeyes]{\textbf{and, sir, Dr}}
& S1, +[MASK] + S2:

\hlc[pink]{\textbf{then, afterwards, till}}

\\
 \hline
COPA   & \underline{Premise}: The girl found a bug in her cereal.

\underline{Positive Hypothesis}: She lost her appetite.

\underline{Negative Hypothesis}: She poured milk in the bowl. & Pre+ [MASK] + Hypo:  

\hlc[babyblueeyes]{\textbf{P: then, but, and}}

\hlc[babyblueeyes]{\textbf{N: then, next, so}}
& Pre + [MASK] + Hypo: 

\hlc[pink]{\textbf{P: so, therefore, thus}}.

\hlc[pink]{\textbf{N: but, instead, and}}.
\\
\bottomrule
\end{tabular}
\caption{\label{tab:casestudy}Examples from evaluation datasets. Connectives in blue are predicted by the \texttt{BERT-large} model and ones in pink are predicted by \name~(\texttt{BERT-large}). ``P'' and ``N'' represent the positive and negative candidates.}  
\end{table*}

To further verify the effectiveness of proposed methods, we compare with the baseline that the \texttt{BERT-large} models are fine-tuned with only token-level MLM objective on syntactic eventuality sequences.
The performance dropped close to finetuning over original \texttt{BERT-large}, which shows that the gains of \textsc{CoCo}LM are not simply from the MLM objective and the new proposed objectives as well as masking strategies contributed largely.

\subsection{Effect of Different Knowledge Resources}

To access the effect of high-order \textsc{ASER} commonsense knowledge, we compare with the performance of directly integrating single-hop edges from \textsc{ASER}.
We decompose the multi-hop sequences into single-hop edges and keep the comparable size of single-hop edges with multi-hop ones.
The results are shown in Table~\ref{tab:ablation_resource}, from which we can see that there is still a notable gap between multi-hop and single-hop knowledge injection at the comparable size. 
Hence multi-hop knowledge is crucial for LMs to understand eventualities.
Besides \textsc{ASER}, another important event knowledge resource is ATOMIC~\cite{sap2019atomic}, which is a crowdsourced commonsense knowledge graph that contains nine types of \textit{if-then} casual relations between social-centric events. 
However, it is a bipartite graph, which symbolically random walk over ATOMIC is impossible like normal graphs.  
Nevertheless, we are interested in the differences of injecting human-annotated and auto-extracted triplets into LMs.
Though relation types and triplet size\footnote{The detailed comparison is included in the Appendix\ref{ap:atomic_and_aser}} may vary from other other, \citet{Fang2021DISCOSBT} successfully converts discourse knowledge in \textsc{ASER} to \textit{if-then} knowledge in ATOMIC and shows the former might roughly cover the latter. 
When injecting into LMs, we can see in Table~\ref{tab:ablation_resource} that ATOMIC can outperform the single-hop version of \textsc{ASER} since ATOMIC is cleaner with human annotations. 
We leave how to combine ASER and other event knowledge resources~\cite{mostafazadeh2020glucose,Hwang2020COMETATOMIC2O} to get more high-quality multi-hop event knowledge as our future work.

\subsection{Effects of Knowledge Retrieval}

We also study the effect of other knowledge injection methods, for example simple retrieving relevant nodes from \textsc{ASER}. 
We use the BM25 algorithm to retrieve Top 5 relevant nodes for each event in the ROCStories. 
Following \citet{petroni2020context}, retrieved nodes are appended at the end of story context and separated by the \texttt{[SEP]} token. 
The results show no obvious improvements over baselines. The reason might be that single event nodes could not provide more information and all the tasks require relational knowledge.
Advanced integration methods with retrieved knowledge like \citet{lv-etal-2020-integrating} and \citet{DBLP:conf/icml/GuuLTPC20} are worthy to be deeply explored in the future. 

\subsection{Probing Experiments}

We present one case study from the probing analysis experiment in Table~\ref{tab:casestudy} to further investigate the discourse-aware nature of our proposed language models.
Motivated by~\citet{DBLP:conf/emnlp/PetroniRRLBWM19}, we put a \texttt{[MASK]} token between two events and try to ask the model to predict the connective.
Take the case from COPA dataset as an example, connectives predicted by \name ~clearly show the \textit{effect} relation between two events. However predictions from the baseline models reveal weaker~(\textit{temporal, conjunction}) or wrong~(\textit{contrast}) relations. 
Similar observations could be drawn from another two datasets.
Like Table~\ref{tab:intro_example}, we also sample 300 high-frequency pairs from \textsc{ASER} to predict connectives. 
The P@1 for \name~(\texttt{BERT-large}) has 15.2\% improvement over \texttt{BERT-large}.
These observations show that \name ~manages to memorize richer discourse knowledge about daily events~(Note that connective probing analysis does not mean strong correlations with downstream task performance).



\section{Related Work}\label{sec:related_works}

\noindent \textbf{Understanding Events.} It is important to represent and learn the commonsense knowledge for deeply understanding the causality and correlation between events.
Recently various kinds of tasks requiring multiple dimensional event knowledge are proposed such as story ending prediction~\cite{mostafazadeh-etal-2016-corpus}, event temporal ordering prediction~\cite{ning-etal-2018-joint}, and event casual reasoning~\cite{gordon-etal-2012-semeval}.
Prior studies have incorporated external commonsense knowledge from ConceptNet~\cite{speer2017conceptnet} and ATOMIC~\cite{sap2019atomic} for solving event representation~\cite{ding2019event}, story generation tasks~\cite{guan2020knowledge}, KG completion~\cite{bosselut2019comet}. 
However, their event-level knowledge is sparse and incomplete due to the human-annotated acquisition, which thus limits the model capacity, especially when injecting into LMs. 
\citet{zhang2020aser} builds a large-scale eventuality knowledge graph, \textsc{ASER}, by specifying eventuality relations mined from discourse connectives.
It explicitly provides structural high-order discourse information between events spanning from temporal, casual to co-occurred relations, which has been proven to be transferable to human-defined commonsense~\cite{zhang2020TransOMCS,Fang2021DISCOSBT} and help with script learning~\cite{lv-etal-2020-integrating}.
In this work, we aim at making full use of multi-dimensional high-order event knowledge in the ASER to help pretrained LMs understand events.

\noindent \textbf{Injecting Knowledge into LMs.} 
Though \citet{DBLP:conf/emnlp/PetroniRRLBWM19} shows that pre-trained LMs store factual knowledge without fine-tuning, still, LMs can not handle knowledge-intensive tasks such as open-domain question answering or commonsense reasoning. 
Previous works explore different ways to inject various knowledge into pre-trained LMs for downstream tasks. 
They mainly differ from knowledge resources, masking strategies, and training objectives. 
From the resource side, entity-centric KGs are infused into LMs in the form of linked entities~\cite{zhang2019ernie,peters2019knowledge,xiong2020pretrained}. triplets~\cite{yao2019kg,DBLP:conf/aaai/LiuZ0WJD020,wang2020k} or descriptions~\cite{wang2021kepler,yu2020jaket}.
Besides that, linguistic knowledge (e.g.,synonym/hypernym relations~\cite{lauscher2019specializing}, word-supersense ~\cite{levine-etal-2020-sensebert}, dependency parsing~\cite{wang2020k}, and constituent parsing~\cite{zhou2019limit}) also plays a critical role to improve LMs.
Simple commonsense knowledge from ConceptNet~\cite{speer2017conceptnet} is injected into LMs via linked entity-level MLMs and a new distractor loss function~\cite{shen-etal-2020-exploiting}.
Last but not least, domain-specific knowledge is also customized to improve relevant tasks such as mined sentiment word~\cite{tian-etal-2020-skep}, event temporal patterns~\cite{Zhou2020TemporalCS}, and numerical reasoning data~\cite{ggb2020injecting}.
We refer readers to \citet{safavi-koutra-2021-relational} for the comprehensive survey. 
In this work, we aim at injecting complex commonsense into pre-trained LMs with two significant difference against previous works: 1)~we use the event rather than tokens as the semantic unit, and propose to use an eventuality-based masking strategy as well as two auxiliary tasks to help LMs understand events; 2)~We first leverage the random walk process on a large-scale knowledge graph to include multi-hop knowledge.

\section{Conclusion and Future Work}\label{sec:conclusion}

In this work, we aim at helping pre-trained language models understand complex commonsense about eventualities. 
Specifically, we first conduct the random walk over a large-scale eventuality-based knowledge graph to collect multi-hop event knowledge and then inject the knowledge into the pre-trained LMs with an eventuality-based mask strategy as well as two auxiliary tasks.
Experiments on three downstream tasks as well as extensive analysis demonstrate the effectiveness of the proposed model.
As our approach is a general solution, we believe that it can also be helpful for other tasks that require complex commonsense about events.

For future work, we would sample sub-graph structures to explore more meaningful event-centric commonsense knowledge~\cite{wang2021contextualized}.
Moreover, we will equip our models with generative abilities  
by finetuning powerful T5~\cite{2020t5} or BART~\cite{lewis-etal-2020-bart} models to help narrative story completion~\cite{ji-etal-2020-language}, commonsense inference~\cite{gabriel2021paragraph}, event infilling tasks~\cite{lin-etal-2021-conditional}.
Unified event-aware language models like \citet{zhou2022claret} would be promising and interesting directions.

\section*{Acknowledgements}

Yangqiu Song was supported by the NSFC Fund~(U20B2053) from the NSFC of China, the RIF~(R6020-19 and R6021-20) and the GRF~(16211520) from RGC of Hong Kong, the MHKJFS~(MHP/001/19) from ITC of Hong Kong with special thanks to HKMAAC and CUSBLT, and the Jiangsu Province Science and Technology Collaboration Fund~(BZ2021065).
We would also like to thank Wei Wang for insightful dicussions. 

\bibliography{custom}
\bibliographystyle{acl_natbib}

\appendix

\section{Appendices}
\label{sec:appendix}

\subsection{Examples of Evaluation Datasets}\label{ap:dataset_examples}

We select one example for each commonsense evaluation dataset and list in Table~\ref{tab:dataset_example}. 
In terms of MATRES, it has 13,577 event pairs among 256 articles with 4 temporal relations, i.e., \textit{Before}~(6,874), \textit{After}(4,570), \textit{Equal}~(470) and \textit{Vague}~(1,656).
Compared with MATRES and COPA, solving the ROCStories requires more complex commonsense knowledge to understand the whole narrative and multiple types of relations across events. 

\begin{table}[h]\small
\setlength\tabcolsep{4.0pt}
\centering
\begin{tabular}{p{1.1cm} |p{6.2cm}}
\toprule
\textbf{Dataset} &\textbf{Example}\\
\midrule
ROC
Stories & \underline{Context}: The Mills next door had a new car. The car was stolen during the weekend. They came to my house and asked me if I knew anything. I told them I didn't, but for some reason they suspected me.

\underline{Positive Ending}: They called the police to come to my house.

\underline{Negative Ending}: They liked me a lot after that.
\\ \hline
MATRES  & Fidel Castro \{$e_1$: \underline{invited}\} John Paul to \{$e_2$: \underline{come}\} for a reason. 

\textit{Label}: \textbf{BEFORE} \\
 \hline
COPA  & \underline{Premise}: I knocked on my neighbor's door. 

\underline{Positive Hypothesis}: My neighbor invited me in. 

\underline{Negative Hypothesis}: My neighbor left his house.\\
\bottomrule
\end{tabular}
\vspace{-0.5mm}
\caption{The examples for all commonsense evaluation datasets. \label{tab:dataset_example}}  
\end{table}

\subsection{ASER Discourse Relations}

In the Table~\ref{tab:relation_marker}, we list ten discourse relations as well as representative connectives~(markers) used to train \name. 
We further categorize them into three types: ``temporal'', ``casual'' and ``others''.
We refer the readers to original ASER papers~\cite{zhang2020aser} for detailed relation analysis. 

 \begin{table}[h]\small
 \centering
 \setlength\tabcolsep{5pt}
 \begin{tabular}{l|c|c}
 \toprule
  Types  &  Relations &  Connectives     \\ \midrule 
 \multirow{3}{*}{\begin{minipage}{0.8in}Temporal \end{minipage}}& Precedence  &  before    \\
 & Succession  & after    \\ 
 & Synchronous  & meanwhile     \\ 
 \midrule
 \multirow{3}{*}{\begin{minipage}{0.8in}Casual \end{minipage}}& Reason  & because       \\
 & Result & so      \\
 & Condition  & if  \\ \midrule
 \multirow{4}{*}{\begin{minipage}{0.8in}Others \end{minipage}}& Conjunction  & and       \\
 & Contrast & but      \\
 & Alternative  & or  \\
 & Concession  & although \\
 \bottomrule

 \end{tabular}
 \caption{\label{tab:relation_marker}The discourse relations as well as representative markers in the ASER knowledge graph.}  
 \end{table}
 
 \subsection{Eventuality Sequences}
 
 We append more sampled eventuality sequences from random walk. 
 Also we organize the sequences into several meta-paths~(the paths with same relation patterns).
 Here only 2-hop and 3-hop sequences are listed and we could observe meaningful high-order connections between eventualities. 
 
 \begin{figure}[h]
  \centering
  \includegraphics[width=0.80\columnwidth]{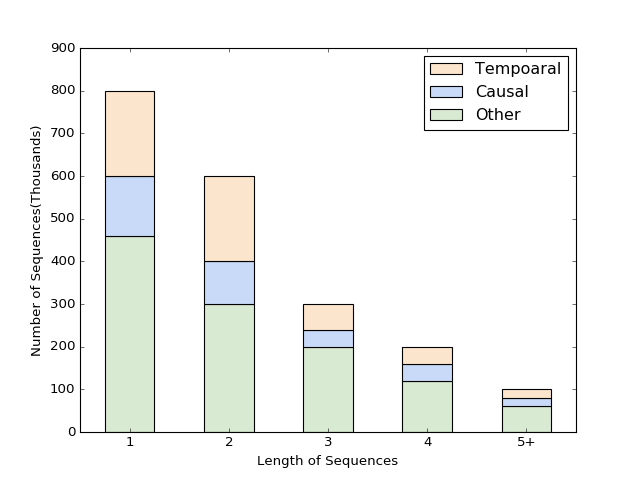}
  \caption{The distribution of lengths along with relation edges for generated eventuality sequences.
  }\label{fig:seq_dis}
  \vspace{-0.4cm}
\end{figure}

The sequence distributions with different lengths and types of relations are shown in the Figure~\ref{fig:seq_dis}.
We can see that casual relations take up a small share, which again show that causal knowledge tends to be implicit and hard to acquire.

\subsection{ATOMIC V.S. ASER}\label{ap:atomic_and_aser}

In this section, we summarize the nine casual/effect relations from ATOMIC~\cite{sap2019atomic} in the Table 2. 
\citet{Fang2021DISCOSBT} shows ASER's discourse relations could be converted to causal knowledge in the ATOMIC.
Thus ASER might roughly cover the knowledge in the ATOMIC.
Moreover, ASER also covers agentless events such as ``the weather is good'', which was partially covered by GLUCOSE~\cite{mostafazadeh2020glucose}
However it contains noise compared with ATOMIC. 
\name experiments show ATOMIC~(877K edges) performs better than ASER(4.4 M - 5$\times$ larger).
 
 \begin{table}[h]\small
 \centering
 \setlength\tabcolsep{3pt}
 \begin{tabular}{l|c|c}
 \toprule
  If-Then Types  &  Relations &  Causal Types     \\ \midrule 
 \multirow{3}{*}{\begin{minipage}{1.2in}If-Event-Then-State \end{minipage}}& xIntent  &  Cause   \\
 & xReact  & Effect    \\ 
 & oReact  & Effect   \\ 
 \midrule
 \multirow{5}{*}{\begin{minipage}{1.2in}If-Event-Then-Event \end{minipage}}& xNeed & Cause \\
 & xEffect  & Effect       \\
 & xWant  & Effect \\
 & oEffect & Effect     \\
 & oWant  & Effect  \\ \midrule
 \multirow{1}{*}{\begin{minipage}{1.2in}If-Event-Then-Persona \end{minipage}}& xAttr  & Stative \\
 \bottomrule

 \end{tabular}
 \caption{\label{tab:atomic_relation_marker}The \textit{if-then} types and causal relations between events in the ATOMIC knowledge graph. For relations, ``x'' and ``o'' refer to PersonX and others while ``xAttr'', ``xIntent'', ``xReact'' mean the attribute, intent, reaction of PersonX \textit{etc.}}  
 \end{table}

\end{document}